%% file: main.tex
\documentclass[10pt,twocolumn,letterpaper]{article}

\makeatletter
\providecommand{\@LN}[2]{}
\makeatother

\usepackage{cvpr}              


\definecolor{cvprblue}{rgb}{0.21,0.49,0.74}
\usepackage[pagebackref,breaklinks,colorlinks,allcolors=cvprblue]{hyperref}

\usepackage{xcolor}
\usepackage{graphicx}
\usepackage{amsmath}
\usepackage{amssymb}
\usepackage{colortbl}
\usepackage{booktabs}
\usepackage{amsmath,amsthm,amssymb,amsfonts}
\usepackage[ruled]{algorithm2e} 
\usepackage{enumitem} 
\usepackage{caption}
\usepackage{pifont}
\usepackage{multirow}
\usepackage{enumerate}
\usepackage{color}

\newcommand{\av}[0]{\ensuremath{\boldsymbol{a}} }

\usepackage[capitalize]{cleveref}
\crefname{section}{Sec.}{Secs.}
\Crefname{section}{Section}{Sections}
\Crefname{table}{Table}{Tables}
\crefname{table}{Tab.}{Tabs.}

\def\paperID{11191} 
\def\confName{CVPR}
\def\confYear{2025}

\title{Discovering Fine-Grained Visual-Concept Relations by Disentangled Optimal Transport Concept Bottleneck Models}

\author{Yan Xie$^{1}$\thanks{Equal contribution. \hspace{4mm}  \textdagger Corresponding authors} \hspace{2mm}
Zequn Zeng$^{1}$\footnotemark[1] \hspace{2mm}
Hao Zhang$^{1}$\footnotemark[2] \\
Yucheng Ding$^{1}$ \hspace{2mm}
Yi Wang$^{1}$ \hspace{2mm}
Zhengjue Wang$^{2}$ \hspace{2mm}
Bo Chen$^{1}$ \hspace{2mm}
Hongwei Liu$^{1}$\\
$^{1}$National Key Laboratory of Radar Signal Processing, Xidian University, Xi’an, 710071, China\\
$^{2}$State Key Laboratory of Integrated Service Networks, Xidian University, Xi’an, 710071, China\\
{\tt\small\{yanxie0904, zzequn99, zhanghao\_xidian\}@163.com, bchen@mail.xidian.edu.cn}}

\definecolor{mygreen}{RGB}{34,139,34}

\begin{document}

\maketitle

\begin{abstract}
Concept Bottleneck Models (CBMs) try to make the decision-making process transparent by exploring an intermediate concept space between the input image and the output prediction.
Existing CBMs just learn coarse-grained relations between the whole image and the concepts, less considering local image information, leading to two main drawbacks:
i) they often produce spurious visual-concept relations, hence decreasing model reliability;
and ii) though CBMs could explain the importance of every concept to the final prediction, it is still challenging to tell which visual region produces the prediction.
To solve these problems, this paper proposes a Disentangled Optimal Transport CBM (DOT-CBM) framework to explore fine-grained visual-concept relations between local image patches and concepts.
Specifically, 
we model the concept prediction process as a transportation problem between the patches and concepts, thereby achieving explicit fine-grained feature alignment. We also incorporate orthogonal projection losses within the modality to enhance local feature disentanglement. To further address the shortcut issues caused by statistical biases in the data, we utilize the visual saliency map and concept label statistics as transportation priors. 
Thus, DOT-CBM can visualize inversion heatmaps, provide more reliable concept predictions, and produce more accurate class predictions.
Comprehensive experiments demonstrate that our proposed DOT-CBM achieves SOTA performance on several tasks, including image classification, local part detection and out-of-distribution generalization.

\end{abstract}

\section{Introduction}
\label{sec:intro}
Explainable artificial intelligence aims to allow human users to comprehend and trust the output created by machine learning algorithms~\cite{huysmans2011empirical,christoph2020interpretable}.
Compared with post-hoc explanation methods~\cite{selvaraju2017grad} for black-box models, designing explicit interpretability during modeling and learning processes is beneficial to obtaining more natural explanations~\cite{de2013decision}.

\begin{figure}[t!]
  \centering
    \begin{subfigure}{0.9\linewidth}
        \includegraphics[width=1\linewidth]{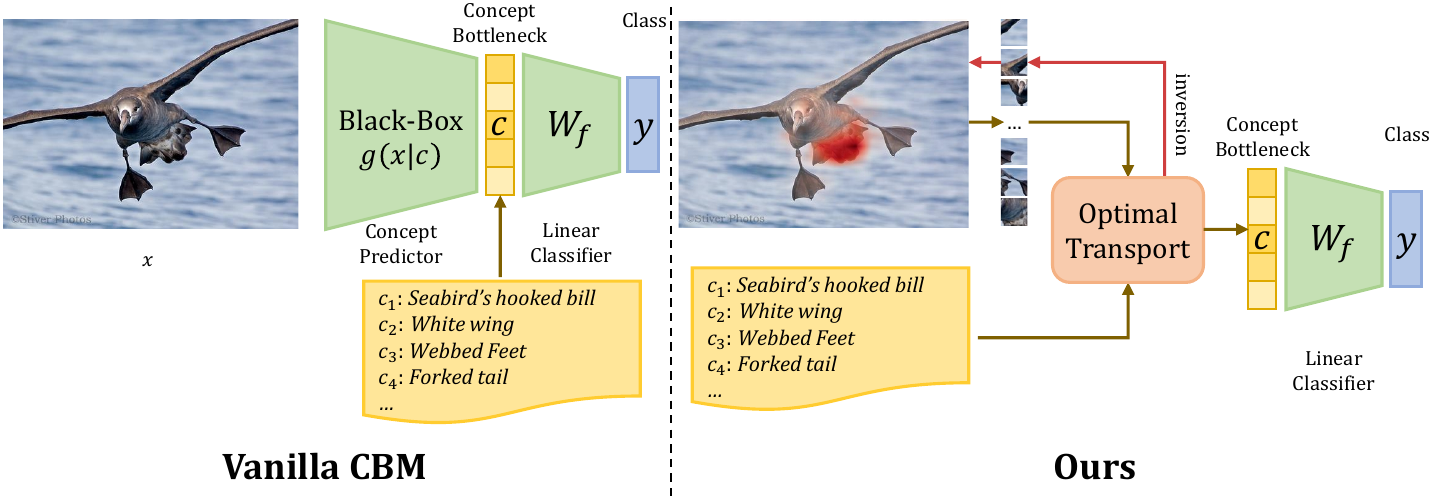}
        \caption{Brief comparison about the coarse-grained design of vanilla CBM and fine-grained design of our proposed DOT-CBM. }
        \label{fig:1_a}
    \end{subfigure}
    
    \vspace{1em} 
    
    \begin{subfigure}{0.9\linewidth}
        \includegraphics[width=1\linewidth]{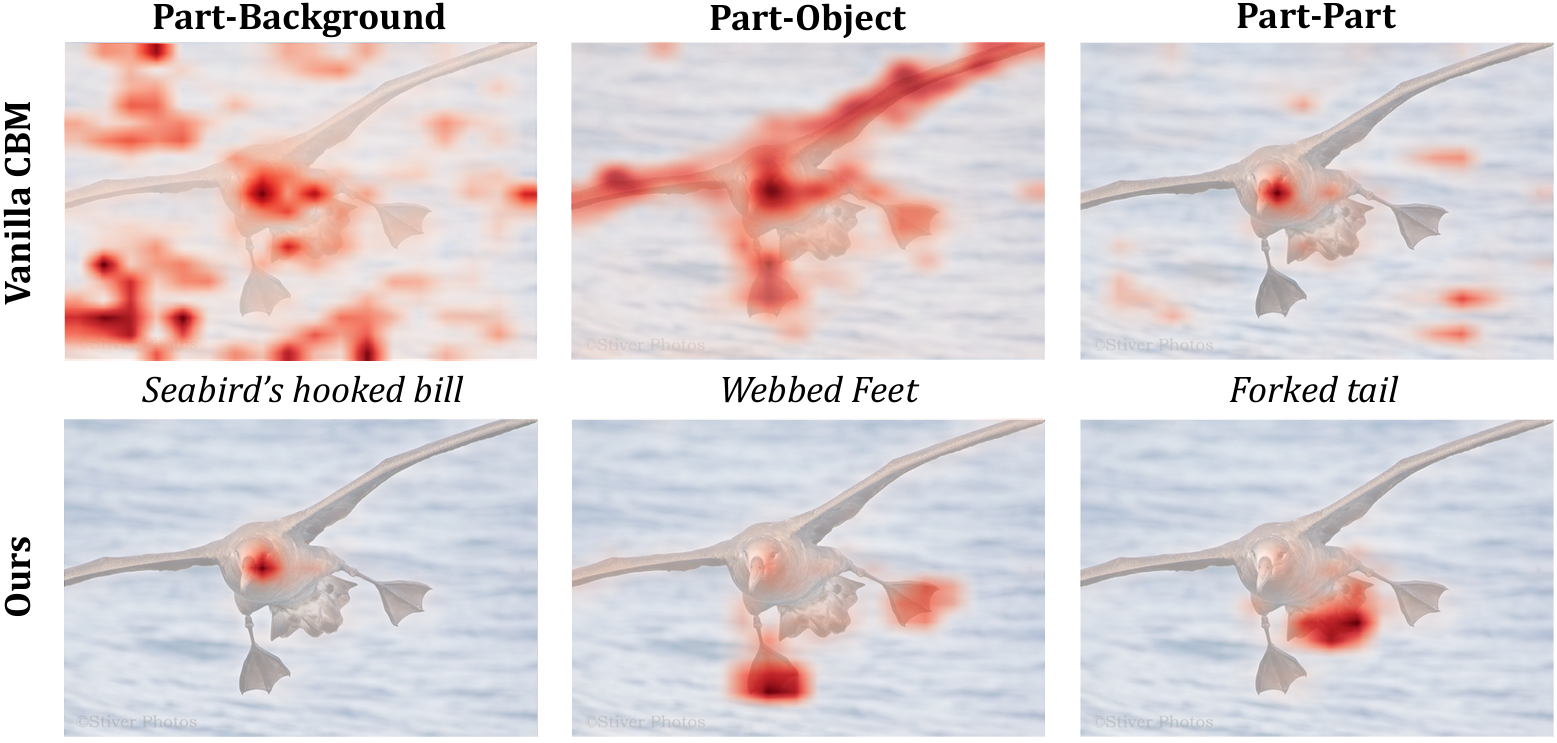}
        \caption{Inversion heatmap visualization. }
        \label{fig:three-level}
    \end{subfigure}
    \vspace{-3mm}
    \caption{Comparison between vanilla CBM and our proposed DOT-CBM, including architecture design and inversion heatmap for concept predictions. (a) Due to the black-box mapping from images to concepts, Vanilla CBM needs Grad-CAM~\cite{selvaraju2017grad} techniques to locate concept predictions back to image space while DOT-CBM can provide an explicit inversion heatmap visualization. (b) Due to lack of fine-grained alignment, vanilla CBM produces spurious correlations in three levels of granularity (part-background, part-object and part-part level) that mislocalizes the local concept to the background, the whole object, and incorrect local region.}
    \label{fig:main}
\vspace{-3mm}
\end{figure}

The Concept Bottleneck Model (CBM)~\cite{koh2020concept}, a representative interpretable image classification model, first predicts a pre-defined set of concepts from images and then maps these concept descriptions to the final output class.
Therefore, CBM is good at interpreting the relationship between the input image and output class predictions via intermediate human-understandable concepts. 
However, vanilla CBM brings these good properties at the expense of classification performance~\cite{midavainere}.
For this problem, a series of CBM variants are proposed from two lines, construct a more comprehensive and discriminative concept bank~\cite{yuksekgonul2022post,yang2023language,oikarinen2023label}, and sophisticated architecture design~\cite{Moayeri_2023_CVPR, semenov2024sparse, sheth2024auxiliary} for better concept representation learning.


Although these variants narrow the performance gap between CBMs and their black-box counterparts, their modeling method between images and concepts still has two disadvantages, as shown in Fig.~\ref{fig:1_a}.
Firstly, they predict multiple local concepts ($\ie$ local attributes like ``head'') from a global image feature, which only learns a coarse-grained relationship between the full image and all concepts.
Secondly, the interpretability of prevailing CBMs mainly comes from the linear concept-class relations. 
The mapping from the original image to all concepts is still modeled as the black-box neural networks, which still need some post-hoc analysis~\cite{margeloiu2021concept, midavainere} to visualize the learned visual-concept relations.
As shown in Fig.~\ref{fig:three-level}, these two disadvantages often bring some visual-concept spurious correlations, where many concept predictions are frequently mis-localized in the images. These spurious correlations inherently undermine the interpretability and generalizing capabilities of CBMs.

Empirically, we categorized these visual-concept spurious correlations into three levels of granularity: \textit{i) part-background spurious correlation.} This refers to a misattribution of the concept predictions of local parts to the background. \textit{ii) part-object spurious correlation.} The concepts of local parts are wrongly associated with the entire object without further distinctions. \textit{iii) part-part spurious correlation.} In this case, the local concepts are mislocalized to incorrect image regions. We attribute these spurious correlation problems to two key factors. The first one is the lack of fine-grained alignment between the local image regions and corresponding textual concepts, $\ie$ the fine-grained visual-concept relations. The second one is the overfitting of the data bias such as the statistically high co-occurrence between class and background and between concept and concept, yielding confusion to learn the difference between them.  
Besides, it is intuitive that the explicit interpretability modeling for the concept predictions from the images is conceivably effective in addressing these visual-concept spurious correlations.

To this end, we propose a novel \textbf{D}isentangled \textbf{O}ptimal \textbf{T}ransport \textbf{C}oncept \textbf{B}ottleneck \textbf{M}odels, namely \textbf{DOT-CBM}. Compared with current CBMs, our proposed DOT-CBM can provide holistic explicit interpretability on not only the concept-class relations but also the visual region-concept relations. 
Specifically, we introduce a disentangled optimal transport (DOT) framework to ensure the intra-modality disentanglement and inter-modality alignment between the image patches and textual concepts. 
The transport assignment solution of OT framework, is an explicit correlation matrix between image patches and pre-defined concepts, facilitating bidirectional outputs including concept predictions on concept space and concept inversion mask on the image space.
Furthermore, to tackle the overfitting to the data bias, we introduce two discrete prior distributions on the patch set and the concept set to penalize the incorrect shortcut alignment between patches and concepts.



In summary, our contributions are as follows:
\begin{itemize}
\item We empirically investigate the visual-concept spurious correlation phenomenon and pinpoint the two key factors of these limitations, $\ie$ lack of fine-grained visual-concept alignment and overfitting to the data bias.

\item We propose a novel disentangled optimal transport framework to model the fine-grained visual-concept alignment and address the overfitting problems.
The transport design between image patches and textual concepts provides explicit relations between image and concept predictions, which can produce concept predictions on the concept spaces and concept inversion mask on the image space.

\item Extensive experiments on image classification, part detection, and out-of-distribution generalization benchmarks demonstrate our proposed DOT-CBM can achieve significant improvements over current state-of-the-art CBMs on classification accuracy, interpretability and generalizing capabilities to new distributions. 
    
\end{itemize}


\begin{figure*}
  \centering
  \includegraphics[width=1\linewidth]{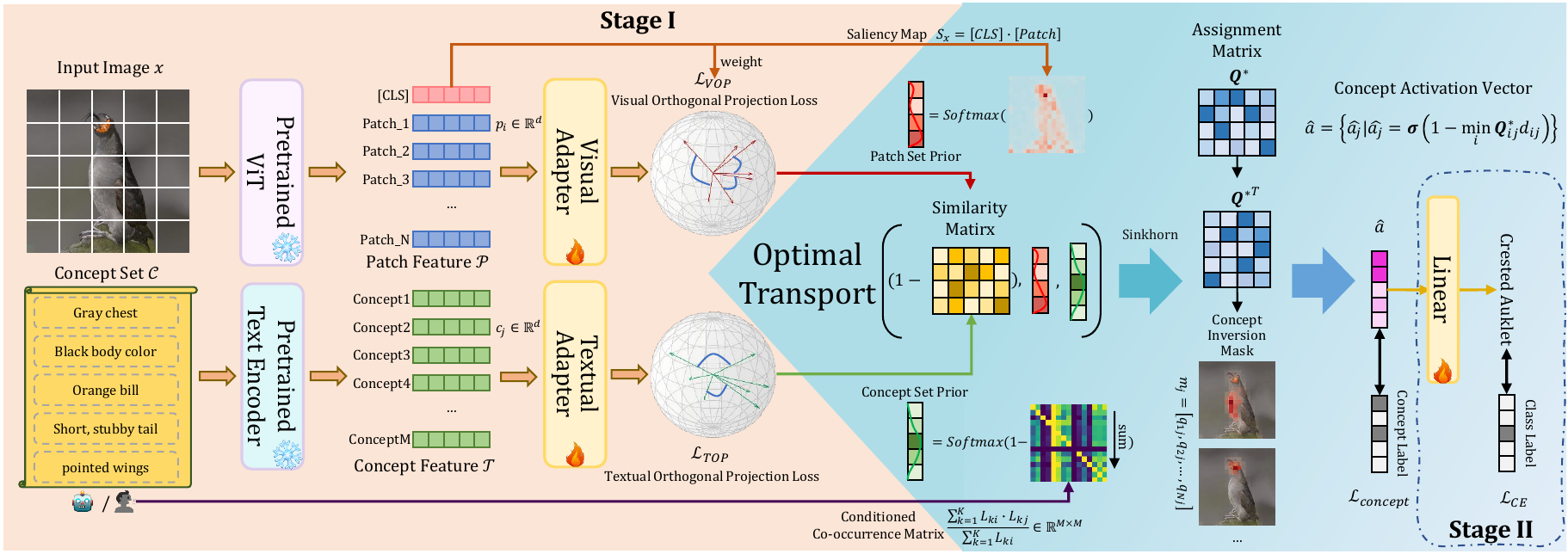}
  \vspace{-5mm}
  \caption{Overview of our proposed DOT-CBM. The overall CBM framework proceeds from left to right. In the first stage, the model transforms input images into concept activation vectors. In forward pass, local embeddings from a pre-trained Vision Transformer (ViT) and a text encoder are processed through learnable adapters to generate two feature sets (Sec.~\ref{local}). Two orthogonal projection losses are applied to both feature sets to constrain Adapter training. An Optimal Transport (OT) algorithm optimizes the Assignment Matrix, representing the explicit correlation between concepts and image patches. Concept activation values are derived by combining this matrix with a cost matrix, and the loss is supervised by concept labels for training (Sec.~\ref{ot}).To address data bias, we use the Saliency Map from the pre-trained ViT for rough foreground-background separation as the visual prior in OT. A Conditioned Co-occurrence Matrix, based on frequency statistics, serves as the prior for concepts, enhancing the model's ability to distinguish co-occurring concepts (Sec.~\ref{prior}).  In the second stage, consistent with the vanilla CBM framework, concept activation values are used to predict final class labels via a linear classification network, which is trained using class labels (Sec.~\ref{overall}).}
  \label{fig:framework}
\end{figure*}

\section{Related Work}

\textbf{Concept Bottleneck Models} 
are a family of models~\cite{koh2020concept} that first predict concepts from input images, then use these concepts to predict a downstream label. 
This design allows CBMs to (1) provide concept-based explanations via their predicted concepts, and (2) improve their test performance when deployed with experts via concept interventions \cite{chauhan2023interactive,shin2023closer,espinosa2024learning}. 
Similar to previous interpretable models, CBMs also face the issue of balancing task accuracy and interoperability~\cite{midavainere,espinosa2022concept}. 
Recent works~\cite{yuksekgonul2022post,oikarinen2023label} has leveraged extra knowledge graph~\cite{yuksekgonul2022post,semenov2024sparse}, expert rules~\cite{konstantinov2024incorporating} or Large Language Models (LLMs)~\cite{yang2023language,semenov2024sparse,li2025deal,oikarinen2023label,zang2024pre} to overcome the limitations of manually defined concept sets and has utilized pre-trained Vision-Language Models (VLMs)~\cite{yang2023language, zeng2023conzic, zeng2024meacap, wen2025beyond, ma2024mode} to obtain concept predictions. 
These advancements have elevated the performance of CBMs to levels approaching those of black-box models.

The interpretability of prevailing CBMs primarily arises from linear relationships between concepts and classes. However, the mapping from the image to the concepts is still handled by black-box neural networks, requiring post-hoc analysis for visualization of the learned visual-concept relations. This lack of transparency often leads to visual-concept spurious correlations, where concept predictions are frequently mis-localized in the image. We propose a disentangled OT to address these limitations.

\textbf{Optimal Transport} aims to find the most efficient way to map one distribution onto another by minimizing a transportation cost.  In the context of machine learning, OT is widely applied in a variety of domains, including unsupervised domain adaptation~\cite{redko2019optimal}, label representation~\cite{frogner2015learning,zhao2018label}, cross-modal semantics~\cite{lin2024multi,lee2019hierarchical,mahajan2019joint,chen2020graph,li2023patchct,chen2022plot}, and beyond. 

In this work, we further extend OT to a disentangled OT framework, to enforce the fine-grained alignment between image patches and their corresponding concepts while also ensuring disentanglement within each modality.




\section{Method}

In this section, we begin with a brief review of the CBM problem setup in Sec.~\ref{problem} including two stages of optimization, $\ie$ patches to concepts ($f_{\mathcal{X} \to \mathcal{C}}$) and concepts to classes ($f_{\mathcal{C} \to \mathcal{Y}}$). 
In the first stage, to address the limited interpretation of previous CBMs in the patches-to-concepts process, we introduce a disentangled optimal transport alignment framework between patches and concepts.
Concretely, in Sec.~\ref{local}, we propose two orthogonal projection losses to disentangle the patch features and concept features.
In Sec.~\ref{ot}, we introduce an optimal transport (OT) based framework to model the complex relationship between patches and concepts, which can predict concepts from patches and meanwhile inverse the concept to patches to provide explanations of concept prediction.
In Sec.~\ref{prior}, we integrate two prior distribution modeling to address the data bias problem.
In the second stage, in Sec.~\ref{overall}, we follow vanilla CBMs and predict classes from concepts. 
Overall, our proposed DOT-CBM achieves holistic interpretability whether on intermediate concept predictions or on final class predictions. The whole pipeline is shown in Fig.~\ref{fig:framework}.

\subsection{Problem Formulation}
\label{problem}
Traditional black-box visual models aim to fit the conditional distribution \( P(y|x) \) given an a dataset \( \mathcal{D} = \{x_i, y_i\}_{i=1}^N \), where $x_i\in\mathcal{X}, y_i\in\mathcal{Y}$ are images and class labels, respectively. Though effective, they still lack transparency to understand the underlying decision-making process to drive the final predictions. 

Concept Bottleneck Models (CBMs)~\cite{koh2020concept} offer an interpretable solution by first predicting human-interpretable high-level concepts before making the final prediction. Formally, the dataset is extended to \( \mathcal{D} = \{x_i, y_i, \av_i\}_{i=1}^N \), where \( \av_i = \{a_{ij}|a_{ij}\in\{0,1\}\}_{j=1}^M \) represents a binary concept activation label. 
 \( a_{ij}=1 \) indicates the presence of a specific high-level concept $c_j$ in image $x_i$, while  \( a_{ij}=0 \) means the absence. $c_j$ is concept from the pre-defined concept set $\mathcal{C}=\{c_j\}_{j=1}^M$.
The optimization process of CBMs can be divided into two-stage~\cite{margeloiu2021concept,raman2023concept}: \textit{1)} they first leverage a concept predictor \( f_{\mathcal{X} \to \mathcal{C}}: \mathbb{R}^{H \times W \times C} \to \mathbb{R}^{M}\) to map the input data \( x \) to the concept activation vector \( \av\in \mathbb{R}^M \). \textit{2)} Subsequently, a target predictor \( f_{\mathcal{C} \to \mathcal{Y}}: \mathbb{R}^{M} \to \mathbb{R}^{N_y} \) maps the concept activation vector \( \av \) to the final label \( y \). This structure enhances the interpretability of the model, as the intermediate concept predictions provide interpretability into the model's final decision. However, the interpretability from image to concept is un-explored.

\subsection{Disentanglement on Local representations}
\label{local}
\textbf{Disentangled visual patch features $\widetilde{\mathcal{P}}$.} As previously discussed, current CBMs typically associate multiple local concepts with a global image feature, which leads to coarse-grained correspondences.  
In this paper, we adopt the pre-trained vision transformer (ViT)~\cite{alexey2020image} model, which has been approved to exhibit better spatial location information extraction than CNN~\cite{raghu2021vision}, to extract local patch features to represent the local image regions.

Specifically, as shown in the left part of Fig.~\ref{fig:framework}, ViT uniformly divides the full image \( x \in \mathbb{R}^{H \times W \times C} \) into $N$ non-overlapping patches $\{I_{pi} \in \mathbb{R}^{P_h \times P_w}\}_{i=1}^N$. After linearly projecting these patches into a $d$-dimension embedding space and flattening all patch embeddings into a sequence, we can get an initialized patch embedding matrix \( X_P \in \mathbb{R}^{N \times d} \). 
To extract the global image feature, ViT concatenates a learnable special token [CLS] with $X_{P}$. Besides, in order to preserve the spatial information of images, ViT adds a learnable position embedding $X_{pos}\in \mathbb{R}^{(N+1) \times d}$ and we get the input embeddings $X$ for ViT:
\begin{align}
    & X_{P} = [I_{p1}\cdot W_p,...,I_{pN}\cdot W_p], \\
    & X = [\text{[CLS]},X_{P}] + X_{pos}, 
\end{align}
where $W_p\in \mathbb{R}^{(P_h \cdot P_w \cdot C) \times d}$ is the patch embedding projection weights.
Then, the pre-trained ViT $E_I$ to map the input embeddings $X$ into the global feature ${g}_I$, and patch features $\mathcal{P}=\{p_i\}_{i=1}^N$:
\begin{align}
[g_I, \mathcal{P}] = E_I(X),
\end{align}

We empirically find that the patch features within the object regions are highly similar which is good for object representation, but bad for distinguishing local parts within the object region.
Thus, we fit a learnable visual adapter $A_v$ on the $\mathcal{P}$ and map the patch features within the foreground region to a disentangled unit hyperspere by introducing a visual orthogonal projection (VOP) regularizer as:
\begin{gather}
S_I = g_I \cdot P, \label{saliency} \\
\langle A_v(p_i), A_v(p_j) \rangle = \frac{A_v(p_i) \cdot A_v(p_j)}{\|A_v(p_i)\| \|A_v(p_j)\|}, \\[2pt]
\mathcal{L}_\text{VOP} = \underset{(p_i,p_j) \sim \mathcal{P}}{\mathbb{E}}[ S_I(i) \cdot S_I(j) \cdot \langle A_v(p_i), A_v(p_j) \rangle] \label{vop}
\end{gather}

where $S_I$ is the saliency map to distinguish foreground and background, $\langle \cdot,\cdot\rangle$ is the cosine similarity. Finally, we get the disentangled patch features $\widetilde{\mathcal{P}}=\{\widetilde{p_i}=A_v(p_i)\}_{i=1}^N$

\noindent\textbf{Disentangled textual concept features $\widetilde{\mathcal{T}}$.}
For concept feature extraction, following ~\cite{Moayeri_2023_CVPR,zang2024pre,yang2023language,li2025deal}, we utilize a pre-trained text encoder $E_T$ to compute concept embeddings $\mathcal{T}=\{t_{j}=E_T(c_j)\}_{j=1}^M$. 
Similarly, we fit a learnable textual adapter $A_t$ and introduce a textual orthogonal projection (TOP) regularizer to disentangle the concept features in a unit hyper-sphere, as:
\begin{align}
\mathcal{L}_\text{TOP} = \underset{(t_i,t_j) \sim \mathcal{T}}{\mathbb{E}}[ \langle A_t(t_i), A_t(t_j) \rangle] \label{top},
\end{align}

Consequently, we get the disentangled concept features $\widetilde{\mathcal{T}}=\{\widetilde{t_{j}}=A_t(t_j)\}_{j=1}^M$.

\subsection{Patch-concept alignment via optimal transport}
\label{ot}
Now, we have the disentangled patch feature set $\widetilde{\mathcal{P}}$ and disentangled concept feature set $\widetilde{\mathcal{T}}$, we need to model the relationship $\mathbf{Q}$ between these two sets and derive the concept activation vector $\hat{\av}$ based on the input image. 
As a complex many-to-many alignment problem, we propose an optimal transport (OT) based binary cross-entropy loss to tackle these problems. Our goal is to correctly align the patch features and concept features with the same semantics.


Assume that these two sets $\widetilde{\mathcal{P}}$ and $\widetilde{\mathcal{T}}$ follow two discrete probability distributions $\boldsymbol{\Theta} \in \mathbb{R}^N$ and $\boldsymbol{\Gamma} \in \mathbb{R}^M$ on the $d$-dimension embedding space, respectively:
\begin{align}
\boldsymbol{\Theta} = \sum_{i=1}^{N} \theta_i \delta_{p_i}, \quad \boldsymbol{\Gamma} = \sum_{j=1}^{M} \gamma_j \delta_{c_j},
\end{align}
where $\boldsymbol{\theta} \in \Sigma^N$ and $\boldsymbol{\gamma} \in \Sigma^M$, the simplex of \(\mathbb{R}^N\) and \(\mathbb{R}^M\), denote the probability values of the discrete states with satisfaction that $\sum_{i=1}^{N} \theta_i = 1$ and $\sum_{j=1}^{M} \gamma_j = 1$.
\(\delta_e\) refers to a point mass located at coordinate \(e\in \mathbb{R}^{d}\). The prior modeling of $\boldsymbol{\theta}$ and $\boldsymbol{\gamma}$ are shown in the Sec.~\ref{prior}.

To get the transport assignment $\mathbf{Q}$ between \(\boldsymbol{\Theta}\) and \(\boldsymbol{\Gamma}\), the objective of OT is formulated as the optimization problem:
\begin{equation}
\label{eq:OT}
        \begin{aligned}
             &\min_{\mathbf{Q}  \in \mathcal{Q}}  \sum_{i,j} q_{ij} d_{ij} - \varepsilon H(\mathbf{Q}) \\
             & \text { s.t. }\mathcal{Q} = \left\{ \mathbf{Q} \in \mathbb{R}_+^{N \times M} \mid  \mathbf{Q} \mathbf{1}_M = \boldsymbol{\theta}, \mathbf{Q}^\top \mathbf{1}_N = \boldsymbol{\gamma} \right\}.
        \end{aligned}
\end{equation}
where $\mathcal{Q}$ is the set of doubly stochastic matrices $\mathbf{Q} = [q_{ij}]_{N \times M}$, denotes the corresponding transport assignment where \(q_{ij}\) represents the probabilities of aligning \( p_i \) with \( t_j \). \(\mathbf{1}^M\) is the \(M\)-dimensional vector of ones. \(d_{ij} = d(p_i, t_j)= (1 - \langle p_i, t_j \rangle)  \geq 0\) is the transport cost between the two points \(p_i\) and \(s_j\) and we use the cosine distance to compute cost matrix $\mathbf{D}$. The optimal $\mathbf{Q}^*$ of \eqref{eq:OT} has a simple normalized exponential matrix solution by Sinkhorn fixed point iterations ~\cite{cuturi2013sinkhorn},

\begin{equation}
\mathbf{Q}^* = \operatorname{Sinkhorn}(\mathbf{D}, \boldsymbol{\theta} ,  \boldsymbol{\gamma} ),
\end{equation}

After obtaining the optimal transport assignment $\mathbf{Q}^*$, the OT distance between the two sets is expressed as:
\begin{equation}
d_{OT} = \sum_{i,j} q_{ij} d_{ij}
\end{equation}

For the concept embedding $t_j$, we only consider the minimum distance $d_{OT_j} = \min_i \left( q_{ij} d_{ij} \right)$ between this concept embedding \( t_j \) and all patch embeddings to facilitate the concept activation value prediction ($x\rightarrow \mathcal{C}$), the predicted concept activation vector $\hat{\av}$ are stated as:
\begin{equation}
    \hat{\av} = \{\hat{a_j}|\hat{a_j}=\boldsymbol{\sigma}(1 - d_{OT_j})\}_{j=1}^M. \label{activation}
\end{equation}
where $\boldsymbol{\sigma}$ is the sigmoid function.
Moreover, we can inverse specific concepts to the image patches based on $\mathbf{Q^*}$, and get the inversion mask $m_j$ of concept $c_j$, as:
\begin{equation}
    m_j = [q_{1j}, q_{2j}, ..., q_{Nj}]. \label{inverse}
\end{equation}
By utilizing OT distance between patches and concepts as the patch-concept transport, our OT-based binary cross-entropy loss $\mathcal{L}_{\text{concept}}$ are as follows:
\begin{equation}
\label{concept}
\mathcal{L}_{\text{concept}} = - \mathbb{E} [ \av \log(\hat{\av}) + (1 - \av) \log(1 - \hat{\av})]
\end{equation}
Now, we can get the total objective of stage 1 ($f_{\mathcal{X}\rightarrow \mathcal{C}}$), as:
\begin{equation}
\label{concept}
\mathcal{L}_{f_{\mathcal{X}\rightarrow \mathcal{C}}} = \lambda_1\mathcal{L}_\text{VOP} + \lambda_2\mathcal{L}_\text{TOP} + \lambda_3\mathcal{L}_\text{concept},
\end{equation}
where $\lambda_1,\lambda_2,\lambda_3$ are the hyperparameters.



The obtained assignment $\mathbf{Q^*}=[q_{i,j}]_{N\times M}$ indicates the relationship between patches and concepts, where $q_{i,j}$ represent correlation between patch $p_i$ and concept $c_j$.
Consequently, based on $\mathbf{Q^*}$, we can obtain concept activation $\av$ (Eq.\eqref{activation}) based on the image features ($x \rightarrow \mathcal{C}$) and meanwhile inverse the concept to the original image patches (Eq.\eqref{inverse}) ($\mathcal{C} \rightarrow x$) to provide explicit interpretation. 

\begin{figure}[t!]
  \centering
\includegraphics[width=0.7\linewidth]{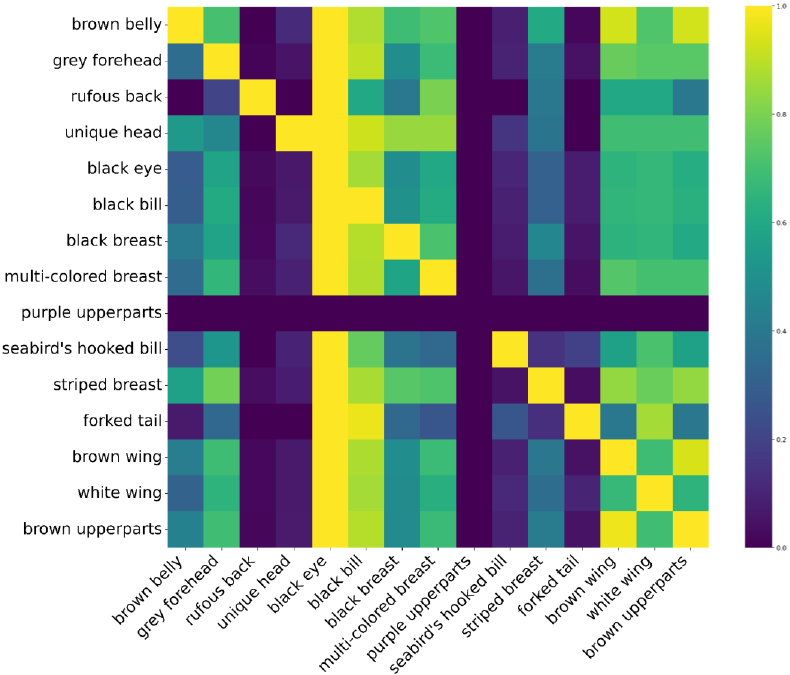}
\caption{Conditioned co-occurrence matrix of concept labels.}
\vspace{-3mm}
\label{fig:occur}
\end{figure}

\subsection{Prior distributions Modeling}
\label{prior}
The prior distributions of $\boldsymbol{\theta}$ and $\boldsymbol{\gamma}$ of OT are typically uniform distributions, which is sub-optimal for our task.
Considering the part-background spurious correlation and part-part spurious correlation problem mentioned in Fig.~\ref{fig:three-level}, we propose two prior distribution modeling in this section to address these limitations, respectively.

\noindent\textbf{Patch set prior $\boldsymbol{\theta}$.} 
We attribute the part-background spurious correlation phenomenon to the high co-occurrence of some classes and backgrounds, such as ``seabird'' and ``water''. 
To solve this biased association between object concepts and the background, we use saliency map $S_I$ from Eq.~\eqref{saliency} to encourage the transports between the patch features within foreground regions and concepts and penalize the transports of background regions.
Specifically, the prior distribution of patch set $\boldsymbol{\theta}$ are formulated as:
\begin{align}
 \boldsymbol{\theta} = \text{Softmax}(S_I) ,
\end{align}





\noindent\textbf{Concept set prior $\boldsymbol{\gamma}$.}
As illustrated in Fig.~\ref{fig:occur}, some concepts are highly co-occurrent. These concept label biases lead to the model's confusion between these concepts, yielding the part-part spurious correlation phenomenon. To address this problem, we employ the statistical co-occurrence matrix of the training dataset to facilitate the prior distribution modeling of $\boldsymbol{\gamma}$, aiming to penalize those concepts with high co-occurrence with other concepts.
Specifically, for a dataset with \( M \) concepts and \( K \) classes, class-concept label form a matrix $L=[l_{i,j}]_{K\times M}$, where \( l_{i,j} \) represents the activation value of the \( j \)-th concept in the concept label for the \( i \)-th class.
First, we count the co-occurrence frequencies of each pair of concepts. Then, we normalize these counts by the total number of times each concept is activated, resulting in the conditional co-occurrence matrix.
The conditional co-occurrence vector \( r_i \) is calculated as follows:
\begin{equation}
r_i = \sum_{j=1}^{M} \left( \frac{\sum_{k=1}^{K} L_{ki} \cdot L_{kj}}{\sum_{k=1}^{K} L_{ki}} \right)
\end{equation}
The concept set prior distribution \( \boldsymbol{\gamma} \) is then derived from the conditional co-occurrence vector as follows:
\begin{equation}
\boldsymbol{\gamma} = 1 - \text{Softmax}(r_i)
\end{equation}

This prior modeling ensures that those concepts with high co-occurrence with other concepts will be penalized.

\subsection{Concept $\rightarrow$ class prediction}
\label{overall}
Now, we already finished the first stage which obtains the intermediate concept activation predictions $\av$. In the second stage, our goal is to learn the association between concept and final class prediction similar to vanilla CBM. Specifically, we can map the $\av$ to final output $\hat{y}$, as:
\begin{align}
\hat{y} = W_F\cdot \av,
\end{align}

The classification objective of stage 2 ($f_{\mathcal{C}\rightarrow \mathcal{Y}}$) is the cross-entropy loss $\mathcal{L}_\text{CE}(y,\hat{y})$. 

\begin{table*}[!htpb]
\centering
\begin{tabular}{l|cccc|cccc}
\toprule
\multirow{2}{*}{Method} & \multicolumn{4}{c|}{Classification Accuracy ($\uparrow$)} & \multicolumn{4}{c}{Part Detection $mAP_{0.5}$ ($\uparrow$)} \\
& ImageNet & CUB & CIFAR100 & AWA2 & PartImageNet & CUB & RIVAL & PASCAL-Parts  \\
\midrule
Vanilla-CBM~\cite{koh2020concept} & 79.17 & 78.32 & 80.04 & 93.15 & 26.94 & 27.02 & 26.76 & 17.32  \\
CEM~\cite{zarlenga2022concept} & 81.29 & 80.47 & 81.23 & 95.92 & 29.19 & 30.86 & 30.83 & 20.85  \\
LaBo~\cite{yang2023language}  & 82.93 & 81.30 & 84.10 & 96.92 & 38.27 & 39.28 & 40.34 &30.17  \\
SparseCBM~\cite{semenov2024sparse} & 82.85 & 82.07 & 84.75 & 95.56  & 39.84 & 40.23 & 40.68 & 33.71 \\
CoopCBM~\cite{sheth2024auxiliary} & 82.73 & 82.10 & 84.66 & \textbf{97.08} & 41.17 & 45.16 & 43.82 & 35.75 \\
\midrule

\textbf{DOT-CBM} & \textbf{83.84} & \textbf{85.39} & \textbf{85.83} & 96.83 & \textbf{50.12} & \textbf{53.47} & \textbf{50.93} & \textbf{44.18}  \\
  & \textcolor{mygreen}{(+0.91)} & \textcolor{mygreen}{(+3.29)} & \textcolor{mygreen}{(+1.08)} & \textcolor{red}{(-0.25)} & \textcolor{mygreen}{(+8.95)} & \textcolor{mygreen}{(+8.31)} & \textcolor{mygreen}{(+7.11)} & \textcolor{mygreen}{(+8.43)}  \\
\bottomrule
\end{tabular}
  \vspace{-3mm}
\caption{Performance comparison of Classification Accuracy (\%) and Part Detection ($mAP_{0.5}$).}
\label{tab:main result}
  \vspace{-3mm}
\end{table*}

\begin{table}[!tb]
\centering
\setlength{\tabcolsep}{3pt}
\begin{tabular}{l|cc|cc}
\toprule
\multirow{2}{*}{Method} & \multicolumn{2}{c|}{CUB} & \multicolumn{2}{c}{Dogs} \\
& ID & OOD & ID & OOD \\
\midrule
Vanilla-CBM~\cite{koh2020concept} & 86.9 & 27.7 & 87.3 & 29.4 \\
CEM~\cite{zarlenga2022concept} & 84.0 & 34.0 & 83.8 & 36.5 \\
LaBo~\cite{yang2023language} & 85.3 & 39.4 & 85.9 & 41.7 \\
Sparse-CBM~\cite{semenov2024sparse} & 84.5 & 35.8 & 85.8 & 38.6 \\
Coop-CBM~\cite{sheth2024auxiliary} & 85.8 & 36.2 & 86.1 & 40.3 \\\midrule
\textbf{DOT-CBM w/o. prior} & 86.5 & 36.9 & 87.0 & 40.8 \\
\textbf{DOT-CBM w. prior} & \textbf{87.2} & \textbf{49.7} & \textbf{89.1} & \textbf{52.4} \\
 & \textcolor{mygreen}{(+0.3)} & \textcolor{mygreen}{(+10.3)} & \textcolor{mygreen}{(+1.8)} & \textcolor{mygreen}{(+10.7)} \\
\bottomrule
\end{tabular}
  \vspace{-3mm}
\caption{Performance comparison of OOD generalization.}
\label{tab:Domain Shift}
  \vspace{-4mm}
\end{table}

\section{Experiment}
To evaluate the effectiveness of our proposed DOT-CBM, we conduct experiments on the image classification task and part detection task.
To further demonstrate our method's abilities to reduce spurious correlations, we also conduct experiments on out-of-distribution generalization tasks. 
For the concept-to-class linear layer, we omit explicit sparse regularization (e.g., L1 penalties) during training. Despite this, the model spontaneously exhibits significant sparsity in the weight matrix due to two orthogonal losses. 
Comprehensive training details, training efficiency results, quantitative evaluations of sparsity and additional experiments are listed in the Appendix.

\subsection{Datasets and Metrics}

\textbf{For the image classification task.} We evaluate our method on the CUB~\cite{wah2011caltech} and AwA2~\cite{xian2018zero} datasets, which are widely adopted in the interpretable community. The CUB dataset contains 11,788 images of 200 bird subcategories, and we use the official train-test split for fine-grained image classification. The AwA2 dataset includes 37,322 images across 50 animal categories, and we follow the same train-test split as in \cite{koh2020concept}. Additionally, we conduct experiments on the larger and more complex ImageNet-1K\cite{deng2009imagenet} and CIFAR-100\cite{krizhevsky2009learning} datasets. 
On these classification datasets, we leverage the classification accuracy to measure the performance of our method.
We also conducted experiments on a more complex dataset (Places365), where the complexity arises from discriminative content being image backgrounds. Quantitative and qualitative results are listed in Appendix.

\noindent\textbf{For the part detection task.} To evaluate the accuracy of the inversion mask produced by DOT-CBM, we evaluate our method on some part detection benchmarks, including CUB dataset, Part-ImageNet~\cite{he2022partimagenet} and PASCAL-Part~\cite{chen2014detect} datasets, which are widely used for part segmentation tasks. We also test our model on the RIVAL-10 dataset~\cite{moayeri2022comprehensive}, which is specifically designed for evaluating model interpretability. 
On the CUB dataset, we obtain the bounding boxes for local parts that are commonly used in part detection tasks\cite{zhang2016spda,xu2020attribute}. The RIVAL-10 dataset consists of over 26K images retrieved from the ImageNet dataset according to the CIFAR-10~\cite{krizhevsky2009learning} categories, and it includes part attribute segmentation masks for the main subjects in the images. The Part-ImageNet and PASCAL-Part datasets are part segmentation datasets that provide fine-grained attribute segmentation masks. We follow ~\cite{morabia2020attention} and preprocess these datasets by focusing on the animal data and merging overly fine-grained part attribute masks. The detailed data preprocessing procedures are described in the Appendix. We use the bounding boxes of these masks as the ground truth of part detection. On these datasets, we use the $mAP_{0.5}$ metric to evaluate the precision of predicted concept inversion masks.

\begin{table*}[!htpb]
\centering
\begin{tabular}{ccc|cccc|cccc}
\toprule
\multirow{2}{*}{OT} & \multirow{2}{*}{Prior} & \multirow{2}{*}{LrD} & \multicolumn{4}{c|}{Classification Accuracy ($\uparrow$)} & \multicolumn{4}{c}{Part Detection $mIOU$ ($\uparrow$)} \\
 & & & ImageNet & CUB & CIFAR100 & AWA2 & PartImageNet & CUB & RIVAL & PASCAL-Parts  \\
\midrule
\ding{55} &\ding{55} &\ding{55} & 81.60 & 80.92 & 82.73 & 96.07 & 0.35 & 0.41 & 0.37 & 0.28  \\
\ding{52} &\ding{55} &\ding{55} & 82.03 & 81.75 & 82.91 & 96.24 & 0.42 & 0.49 & 0.42 & 0.34  \\
\ding{52} &\ding{52} &\ding{55}  & 83.22 & 84.92 & 85.49 & 96.47 & 0.50 & 0.61 & 0.53 & 0.46  \\
\ding{55} &\ding{55} &\ding{52} & 82.73 & 82.10 & 84.66 & 97.08 & 0.47 & 0.58 & 0.49 & 0.41  \\
\ding{52} &\ding{52} &\ding{52} & 83.84 & 85.39 & 85.83 & 96.83 & 0.52 & 0.66 & 0.54 &0.49 \\
\bottomrule
\end{tabular}
\vspace{-3mm}
\caption{Ablation studies on image classification and part detection. OT, Prior, and LrD denote the optimal transport design, prior distribution modeling, and local representation disentanglement.}
\vspace{-2mm}
\label{tab:Ablation_study}

\end{table*}

\subsection{Implementation Details}

We employ the pre-trained DINOv2 ViT-L/14~\cite{oquab2023dinov2} as the image encoder and pre-trained CLIP text encoder~\cite{radford2021learning} as the text encoder. The feature dimension in both encoders is $d = 768$. 
We resize all images to $ H \times W = 252 \times 252 $ as input resolution in both training and testing phases throughout all experiments. For both adapter $A_v$ and $A_t$, we use a three-layer MLP with ReLU activation functions and a dropout rate of 0.4.
The optimization of DOT-CBM is done by AdamW with an initialized learning rate of $1e-4$, and decreased every ten epochs by a factor of 0.5. We set weight decay of 10-5. Batch sizes are set as 128 with maximally 200 training epochs. All experiments are conducted on a single RTX 3090 GPU and our model is implemented in PyTorch. We set $\lambda_1=0.15, \lambda_2=0.2, \lambda_3=0.8 $ for CUB dataset. Hyperparameters setting on other datasets and the range ablation study are in the Appendix.

\subsection{Main Results}
\noindent\textbf{Baselines.} We compare our proposed DOT-CBM with several state-of-the-art CBM baselines from the literature. \textit{Vanilla-CBM}~\cite{koh2020concept} is the first CBM model, proposing the design of the intermediate concept layer. Vanilla-CBM introduces three training strategies, $\ie$ independent, sequential and joint training, in which the independent training can achieve the best interpretation~\cite{margeloiu2021concept}. Our two-stage optimization belongs to the independent training strategy. Additionally, we compare our performance with recent CBM variants including \textit{CEM}~\cite{zarlenga2022concept}, \textit{LaBo}~\cite{yang2023language}, \textit{SparseCBM}~\cite{semenov2024sparse}, \textit{CoopCBM}~\cite{sheth2024auxiliary}. Due to the fact that pre-defined concepts and image backbone will significantly affect the final performance. For a fair comparison, we re-implement all baselines with the high-quality concepts produced by the CDL~\cite{zang2024pre} and a pre-trained ViT backbone to extract image features, the same as our method. The performance of all re-implemented baselines exceeds the best performance reported in their papers.  

\noindent\textbf{Image classification results.} 
As shown in Table ~\ref{tab:main result}, we compare our proposed DOT-CBM with all baselines on four common image classification benchmarks. The classification results show that the DOT-CBM achieves superior performance on three benchmarks. Specifically, the performance improvements over the state-of-the-art (SOTA) are 0.91 points on ImageNet, 3.29 points on CUB, and 1.08 points on CIFAR100. Our method only lags behind CoopCBM by 0.25 points on the AwA2 dataset.

\noindent\textbf{Part detection results.} 
A key property of our proposed DOT-CBM model is able to provide explicit explanations for the concept predictions, via the locate the concept to the original image space, $\ie$ the concept inversion mask derived from the transport assignment. 
This concept inversion mask in the image space indicates the most influential image region contributing to the concept predictions. 
By contrast, all previous CBMs, using a black-box neuron network to model the mapping from global image features to concept predictions, can only inverse the concept prediction to the image space via the post-hoc Grad-CAM.
To evaluate the the accuracy of concept inversion mask, we conduct experiments on the part detection datasets and use the ground-truth bounding boxes of the parts to assess the localization performance. 
To measure the similarity with ground-truths, we need to process our concept inversion mask output and heatmap outputs of baselines (derived from Grad-CAM) into rectangle boxes format. Specifically, we set the threshold at 0.6 of the maximum peak value output of the mask. The local regions above this threshold are considered as the corresponding regions of the concept in the original image. We take the bounding box that encloses this region as the predicted bounding box for the concept. 

From Table.~\ref{tab:main result}, we can observe that DOT-CBM significantly surpasses all baselines (average 8 points improvements), demonstrating the superiority of our design for fine-grained patch-concept alignment.
The Coop-CBM achieves performance that is second only to ours due to the introduction of an orthogonality loss similar to our proposed textual orthogonal project $\mathcal{L}_\text{TOP}$, which further proves the validity of our proposed disentangled OT framework.
Consequently, DOT-CBM can inherently produce accurate and explicit explanations for intermediate concept predictions.
We also conduct comparison with other ViT-appropriate attribution method. Results shown in Appendix.



\subsection{Out-of-distribution generalization}
As discussed in the previous sections, there exists data bias in many datasets, leading to shortcut learning. For example, classifying a dog as a huskie based on the snowy background due to the high co-occurrence in the training set, which is analogous to the part-background spurious correlation discussed in Fig.~\ref{fig:three-level}.
To validate the effectiveness of our method with background shifts, we follow ~\cite{sheth2024auxiliary} to segment the foreground object in CUB\cite{wah2011caltech} and Dogs\cite{KhoslaYaoJayadevaprakashFeiFei_FGVC2011} datasets using SAM~\cite{kirillov2023segment} with bounding box annotations and add a colored background to all images. Each class is associated with a randomly generated color background with an 80\% probability in the training set. The in-distribution (ID) test set contains images with a similar color background probability as the training set, while the color similarity of out-of-distribution (OOD) test set is reduced to 30\%. Results are shown in Table~\ref{tab:Domain Shift}. As we can see, DOT-CBM can achieve comparable results on ID set and surpass the performance of baselines on the OOD set by a large margin, demonstrating that DOT-CBM can effectively decrease the biased association between object concepts and background.
To investigate whether the improvements of tackling part-background spurious correlation are derived from the patch set prior modeling as mentioned in Sec.~\ref{prior}, we conduct ablations for the patch set prior and demonstrate the efficacy of patch set prior $\boldsymbol{\theta}$.

\begin{figure*}
  \centering
  \includegraphics[width=0.9\linewidth]{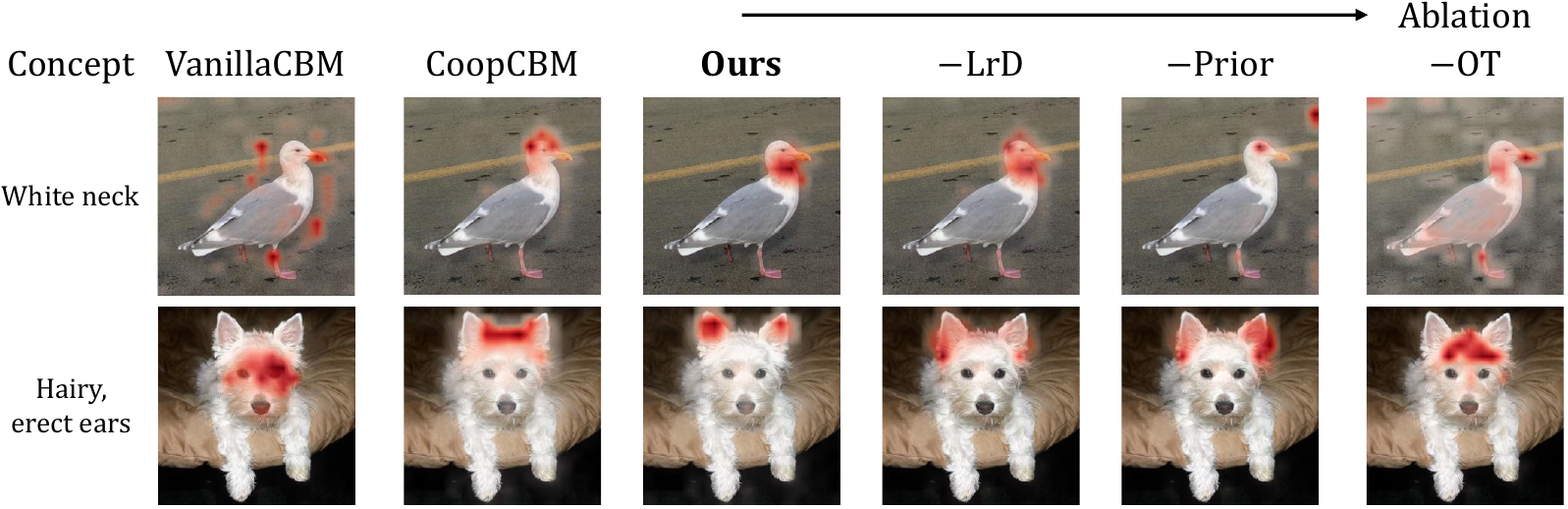}
  \vspace{-3mm}
  \caption{Qualitative results for the concept inversion. VanillaCBM and CoopCBM adopt Grad-CAM to provide a concept inversion heatmap, while our proposed DOT-CBM can generate the inversion mask on its own.}
  \label{fig:qualitative}
  \vspace{-3mm}
\end{figure*}

\subsection{Ablation Studies}

In this section, we investigate the effects of each component in our framework. Ablation results are listed in Table~\ref{tab:Ablation_study}.
The baseline has the same trainable networks as DOT-CBM, including two adapters and a linear classification layer. The baseline replaces the OT assignment with a dot product of patch features and concept features.

\textbf{Effect of OT design (OT).}
For the comparison between the first row and second row in Table.~\ref{tab:Ablation_study}, we can find the incorporation of OT distance, mentioned in Sec.~\ref{ot}, can slightly improve the classification accuracy and largely enhance the part detection performance. OT provides a better distance metric space for modeling the complex many-to-many correlation between concept features and patch features, compared to the direct dot product similarity used in the baseline model.

\textbf{Effect of prior distribution modeling (Prior).} The OT design above, setting the two prior distributions $\boldsymbol{\theta}$ and $\boldsymbol{\gamma}$ as a uniform distribution. 
To mitigate the shortcut learning brought by high occurrence between concept and background and between concept and concept, we add the background penalty patch set prior and the co-occurrence penalty concept set prior, mentioned in Sec.~\ref{prior}. Comparing the second and third rows of Table.~\ref{tab:Ablation_study}, it demonstrates that these two prior distributions can further improve the classification and interpretability.

\textbf{Effect of local representation disentanglement (LrD).} 
After integrating the local representation disentanglement into the baseline (the fourth row of Table.~\ref{tab:Ablation_study}), the improvements are larger than a single incorporation of OT distance (the second row of Table.~\ref{tab:Ablation_study}). The overall DOT-CBM, with a combination of the OT, Prior, and LrD, yields more improvements than individual modules alone.

\textbf{Qualitative results.} Fig.~\ref{fig:qualitative} has shown a qualitative comparison between our method and the baseline methods Vanilla-CBM and CoopCBM.
It is observed that our method is able to more accurately locate the region of the concept in the original image compared to the baseline methods.
Meanwhile, We also present qualitative results by sequentially ablating each component of our framework. When the local representation disentanglement is ablated, the model's localization accuracy decreases to some extent. Further removal of the priors leads to the emergence of part-background spurious correlations. When the OT design is removed, the model produces incorrect localization results.

\vspace{-1mm}
\section{Conclusion.}
\vspace{-1mm}
In this paper, we introduced the Disentangled Optimal Transport Concept Bottleneck Model (DOT-CBM) to address key limitations in existing Concept Bottleneck Models. While CBMs offer interpretable models by mapping image features to human-understandable concepts, they suffer from misalignment between the image and concept, leading to spurious correlations that hinder both interpretability and generalization. 
To overcome these challenges, DOT-CBM introduces a disentangled optimal transport framework that ensures fine-grained alignment between image patches and corresponding textual concepts, thereby providing a more robust and interpretable model. By employing a dual penalty system, we mitigate the overfitting problem caused by data bias.
Empirical results demonstrate that DOT-CBM outperforms previous CBM variants in both classification accuracy and interpretability, offering an effective solution to the problem of visual-concept spurious correlation.

\section{Acknowledgement}
This work was supported in part by the National Natural Science Foundation of China under Grant U21B2006; in part by Shaanxi Youth Innovation Team Project; in part by the Fundamental Research Funds for the Central Universities QTZX24003 and QTZX22160; in part by the 111 Project under Grant B18039;
Hao Zhang acknowledges the support of NSFC (62301384); Excellent Young Scientists Fund (Overseas); Foundation of National Key Laboratory of Radar Signal Processing under Grant JKW202308. Zhengjue Wang acknowledges the support of NSFC (62301407).
{
    \small
    \bibliographystyle{ieeenat_fullname}
    \bibliography{main}
}



\end{document}